\documentclass[journal]{IEEEtran}
\usepackage{}
\usepackage{lineno,hyperref}
\usepackage{graphicx}
\usepackage{epstopdf}
\usepackage{mathrsfs}
\usepackage{amsfonts}
\usepackage{bm}
\usepackage{multirow}
\usepackage{threeparttable}
\usepackage[subfigure]{tocloft}
\usepackage{subfigure}

\ifCLASSINFOpdf
\else
\fi
\begin{document}

\title{Precise Facial Landmark Detection by Reference Heatmap Transformer}

\author{Jun~Wan, Jun~Liu, Jie~Zhou, Zhihui~Lai, Linlin Shen, Hang~Sun, Ping~Xiong, Wenwen Min
	
	
	\thanks{This work is supported by the National Natural Science Foundation of China (Grant No. 62002233, 62076164, 62262069, 61802267, 61976145 and 61806127), the Shenzhen Science and Technology Program (Grant No. JCYJ20210324094601005, JCYJ20210324094413037 and JCYJ20190813100801664), the Natural Science Foundation of Guangdong Province (Grant No. 2021A1515011861) and the Fundamental Research Funds for the Central Universities”, Zhongnan University of Economics and Law (Grant No. 2722023BQ058)). Corresponding author: Zhihui Lai.}
	\thanks{J. Wan is with the School of Information and Safety Engineering, Zhongnan University of Economics and Law, Wuhan 430073, China, and with the Information Systems Technology and Design Pillar, Singapore University of Technology and Design, Singapore, 487372 (e-mail: junwan2014@whu.edu.cn).}
	\thanks{Z. Lai, L. Shen and J. Zhou are with the College of Computer Science and Software Engineering, Shen zhen University, Shenzhen, 518060, China, and the Shenzhen Institute of Artificial Intelligence and Robotics for Society, Shenzhen, 518060, China.(e-mail: lai\_zhi\_hui@163.com, llshen@szu.edu.cn, jie\_jpu@163.com)}	
	\thanks{J. Liu is with the Information Systems Technology and Design
		Pillar, Singapore University of Technology and Design, Singapore, 487372 (e-mail: jun\_liu@sutd.edu.sg).}
	\thanks{H. Sun is with College of Computer and Information Technology, China Three Gorges University, Yichang, HuBei, China. (e-mail: sunhang0418@whu.edu.cn.)}
	\thanks{P. Xiong is  with the School of Information and Safety Engineering, Zhongnan University of Economics and Law, Wuhan 430073, China (e-mail: pingxiong@zuel.edu.cn).}
	\thanks{W. Min is with the School of Information Science and Engineering, Yunnan University, Kunming 650091, Yunnan, China. (e-mail: minwenwen@ynu.edu.cn).}	
}

\markboth{Journal of \LaTeX\ Class Files,~Vol.~14, No.~8, August~2015}%
{Shell \MakeLowercase{\textit{et al.}}: Bare Demo of IEEEtran.cls for IEEE Journals}

\maketitle
\begin{abstract}
Most facial landmark detection methods predict landmarks by mapping the input facial appearance features to landmark heatmaps and have achieved promising results. However, when the face image is suffering from large poses, heavy occlusions and complicated illuminations, they cannot learn discriminative feature representations and effective facial shape constraints, nor can they accurately predict the value of each element in the landmark heatmap, limiting their detection accuracy. To address this problem, we propose a novel Reference Heatmap Transformer (RHT) by introducing reference heatmap information for more precise facial landmark detection. The proposed RHT consists of a Soft Transformation Module (STM) and a Hard Transformation Module (HTM), which can cooperate with each other to encourage the accurate transformation of the reference heatmap information and facial shape constraints. Then, a Multi-Scale Feature Fusion Module (MSFFM) is proposed to fuse the transformed heatmap features and the semantic features learned from the original face images to enhance feature representations for producing more accurate target heatmaps. To the best of our knowledge, this is the first study to explore how to enhance facial landmark detection by transforming the reference heatmap information. The experimental results from challenging benchmark datasets demonstrate that our proposed method outperforms the state-of-the-art methods in the literature.
\end{abstract}

\begin{IEEEkeywords}
Heatmap regression, landmark detection, shape constraint, heavy occlusion, large pose.
\end{IEEEkeywords}

\IEEEpeerreviewmaketitle

\section{Introduction}
\IEEEPARstart{F}{acial} landmark detection (FLD), also known as face alignment, aims to locate the predefined landmarks (e.g., eye corners, nose tip, mouth corners) of a face, and has attracted much attention in the computer vision community. As a hot topic, FLD is crucial for many facial analysis tasks such as face recognition \cite{Huang2020DeepIL, Zhang2022EnhancedGS}, face animation \cite{Ichim2017PhacePF}, facial expression recognition \cite{Wang2020SuppressingUF, Wang2020RegionAN} and 3D face reconstruction \cite{Feng2018Joint3F, Deng2019Accurate3F}. So imprecise landmarks will be propagated to the subsequent tasks and could cause performance deterioration.
\begin{figure}[t]
	\begin{center}
		\includegraphics[width=0.9\linewidth]{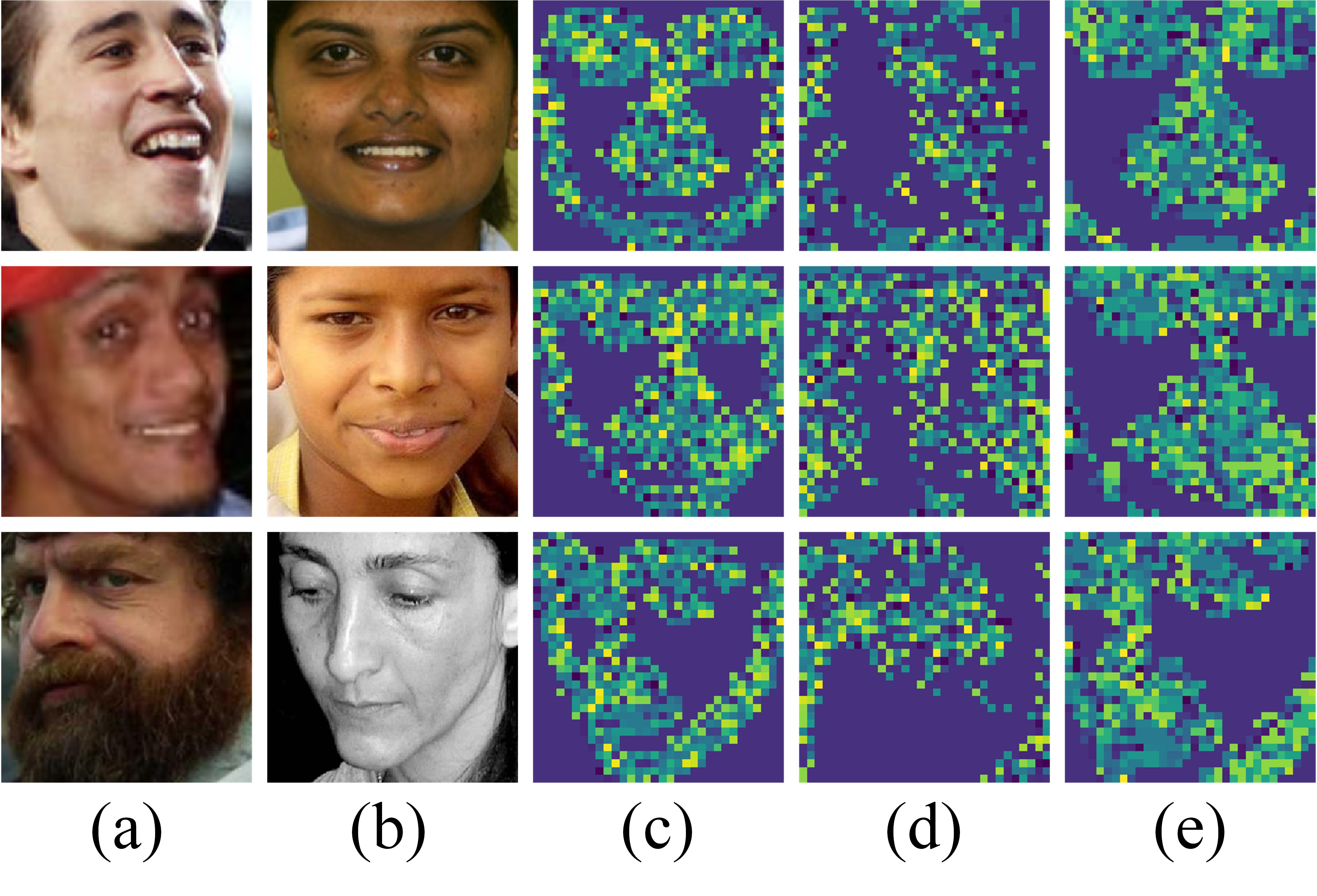}
	\end{center}
	\caption{Visualization of feature maps (The maximum value of all channel elements corresponding to each spatial position is selected). (a) target image. (b) reference image. (c) reference heatmap feature. (d) soft transformation heatmap feature. STM can effectively transfer the correlated reference heatmap information, but at the same time, it also makes the transferred features discrete due to the difference (e.g., face covers, occlusions, etc) between the reference image and the target one. (e) hard transformation heatmap feature. HTM can keep the integrity of the reference heatmap information by transforming it into the target heatmap information with a 2D affine transformation. These two modules can cooperate with each other to encourage the accurate transformation of the reference heatmap information and facial shape constraints.}
	\label{fig1}
\end{figure}

Owning to the development of machine learning \cite{Min2021StructuredSN, Min2021GroupSparseSM, wan2019face, Min2021ANS} and deep learning \cite{wu2020siamese, wan2020robust}, FLD methods \cite{cao2014face, ren2014face, Wu2018LookAB, Dong2018StyleAN, Liu2019SemanticAF, Kumar2020LUVLiFA, Huang2021ADNetLE}, especially heatmap regression-based FLD methods \cite{ Dong2018StyleAN, Liu2019SemanticAF, Kumar2020LUVLiFA, Huang2021ADNetLE, Yang2017StackedHN} have achieved significant improvements. Heatmap regression-based FLD methods usually predict landmark coordinates by producing landmark heatmaps for each landmark, and landmark heatmaps include localization, i.e., a real number (a probability) is supposed to be assigned to each element in the landmark heatmap. Although deep learning-based models have shown strong learning and favorable regression abilities, it is still challenging to accurately estimate the values of all elements in landmark heatmaps and maintain their distribution. Moreover, the occlusions, the large poses, or the imbalances in training samples also mislead Convolutional Neural Network (CNN) on the learning of facial representation and facial shape constraints, which further hinders the generation of landmark heatmaps. Inspired by the great similarity in the structure of human faces, e.g., all faces have two eyes, one mouth, and one nose, the nose is in the middle of the face, and the mouse is always under the nose, this work explores ways to transform and transfer the corresponding landmark heatmaps from one face (i.e., the face with known ground-truth landmark heatmaps) to another one (i.e., the face that need to predict its landmark heatmaps) according to the similarity between them. However, how to estimate the similarity between these two faces and transform the corresponding landmark heatmaps from one face to another one for more precise FLD remains a challenging problem.
\begin{figure*}
	\begin{center}
		\includegraphics[width=0.9\linewidth]{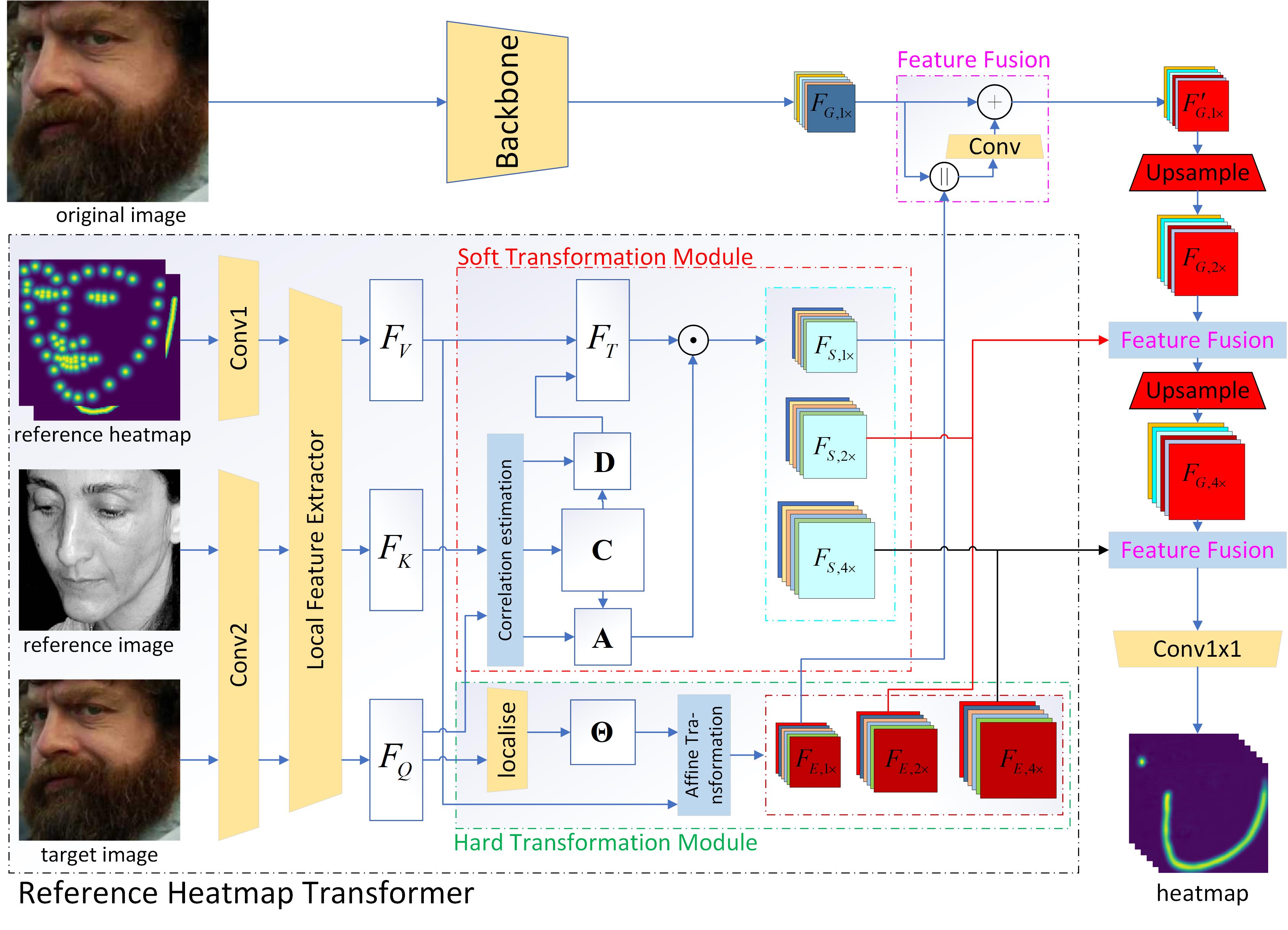}
	\end{center}
	\caption{The overall architecture of the proposed method (i.e., RHT-R). The proposed Reference Heatmap Transformer (RHT) can effectively transform the reference heatmap information into the target ones by integrating with the Soft Transformation Module (STM) and Hard Transformation Module (HTM). Then, the transformed heatmap information will be fused with the semantic features extracted from the original image to produce more accurate target landmark heatmaps.}
	\label{fig2}
\end{figure*}

In this paper, we first name the face with known ground-truth landmark heatmaps as the reference face, and then propose a novel Reference Heatmap Transformer (RHT) to address the above problems by introducing the reference heatmap information, i.e., useful reference heatmap information is selected and transformed by our RHT to help model more effective facial shape constraints and produce more accurate target heatmaps. Specifically, as shown in Fig. \ref{fig2}, RHT consists of a Soft Transformation Module (STM) and a Hard Transformation Module (HTM). The STM and HTM can transform the reference heatmap information in a soft and hard way, respectively. Fig. \ref{fig1} (a) shows the target images that we need to predict their landmark coordinates. Any face image with annotated landmarks can be used as a reference image (as shown in Fig. \ref{fig1} (b)). Of course, an image with a similar head pose is better. The reference heatmap feature (as shown in Fig. \ref{fig1} (c)) is extracted from reference landmark heatmaps. By estimating the correlation between the target image feature and the reference one, STM can dynamically select the correlated reference heatmap features to construct the target heatmap feature (i.e., soft transformation heatmap feature as shown in Fig. \ref{fig1} (d)). At the same time, the HTM constructs the target heatmap feature (i.e., hard transformation heatmap feature as shown in Fig. \ref{fig1} (e)) by transforming the reference heatmap feature with a 2D affine transformation. Finally, the soft and hard transformation heatmap features are combined and fused with the original image's feature (extracted by a mainstream backbone) by a Multi-Scale Feature Fusion Module (MSFFM) to produce target heatmaps. Therefore, by integrating the RHT and MSFFM, more effective facial shape constraints and feature representations are learned to detect more accurate facial landmarks. The contributions of this paper include:

1) By estimating the correlation between the reference image feature and the target one, STM can effectively select and transfer the correlated reference heatmap features into the target ones. 

2) A novel transformer called RHT is proposed to introduce the prior knowledge from reference heatmap information for modeling facial shape constraints and generating landmark heatmaps. To the best of our knowledge, this is the first study to explore how to enhance facial landmark detection by transforming the reference heatmap information.

3) By integrating RHT and MSFFM, more effective facial shape constraints and feature representations can be learned to produce more accurate landmark heatmaps, and our algorithm outperforms state-of-the-art methods on benchmark datasets such as COFW \cite{Burgosartizzu2013Robust}, 300W \cite{Sagonas2016300FI}, AFLW \cite{Zhu2016UnconstrainedFA} and WFLW \cite{Wu2018LookAB}.

The rest of the paper is organized as follows. In Section II, we review related works on facial landmark detection. In Section III, we show the proposed RHT and MSFFM. In Section IV, a series of experiments are conducted to evaluate our proposed method. Finally, we conclude the paper in Section V. 
\section{Related Work}
Research on facial landmark detection can be traced back to the 1990s, and since then, rapid development has been achieved. Early methods for facial landmark detection are based on active shape models \cite{cootes1995active}, active appearance models \cite{cootes2001active}, constrained local model \cite{Cristinacce2006FeatureDA} as well as their variations \cite{Tzimiropoulos2014GaussNewtonDP, Liu2009DiscriminativeFA}, they are sensitive to faces under large poses and partial occlusions. The current FLD methods \cite{ Feng2017WingLF, Wu2018LookAB, Zhu2019RobustFL, Liu2019SemanticAF} are driven by deep convolutional neural networks, and they can be divided into two branches, that are coordinate regression-based and heatmap regression-based.

\textbf{Coordinate Regression-based Methods.} This category of method \cite{cao2014face, ren2014face, zhang2014facial, Trigeorgis2016MnemonicDM, Wu2018LookAB, Zhu2019RobustFL} directly regress landmark coordinate vectors from the input face images. In Mnemonic Descent Method (MDM) \cite{Trigeorgis2016MnemonicDM}, a non-linear unified model is designed to model dependencies between iterations of the cascade by introducing the concept of memory into descent direction learning, thereby facilitating the joint optimization of the regressors and achieving more robust facial landmark detection. Feng et al. \cite{Feng2017WingLF} propose the Wing loss criterion by increasing the contribution of the samples with small and medium size errors to address the data imbalance problem, thereby obtaining more precise landmark detection models. Wu et al. \cite{ Wu2018LookAB} aim to enhance shape constraints by constructing facial boundary heatmaps, and combining them with the original face image features to improve landmark detection accuracy. In cascaded regression and de-occlusion (CRD) \cite{Wan2020RobustFA}, the cascaded deep generative regression model is proposed to address the face de-occlusion and facial landmark detection problems simultaneously, in which these two tasks can be boosted by each other. In occlusion-adaptive deep network (ODN) \cite{Zhu2019RobustFL}, the geometry-aware module, the distillation module and the low-rank learning module are integrated to address the occlusion problem in FLD. However, this kind of method usually regresses landmark coordinates with a fully connected output layer, which ignores the spatial correlations of features and limits their detection accuracy.

\textbf{Heatmap Regression-based Methods.} This kind of method \cite{Yang2017StackedHN, Dong2018StyleAN, Liu2019SemanticAF} predicts landmark coordinates by producing landmark heatmaps. Compared to the coordinate regression-based method, it can effectively keep the original spatial relationships between pixels, thus obtaining promising landmark detection accuracy. Dong et al. \cite{Dong2018StyleAN} propose a style-aggregated approach to address facial landmark detection under style variations, in which the original face images are first transformed to style-aggregated ones and then they are combined to obtain more robust models. Liu et al. \cite{Liu2019SemanticAF} propose a novel latent variable optimization strategy to find the semantically consistent landmark annotations, in which the predicted landmarks become more accurate and the ground-truth shape can also be updated. In MMDN \cite{Wan2021RobustFL} and CCDN \cite{Wan2021RobustFLL}, Wan et al. explore more effective feature representations to achieve more precise facial landmark detection by introducing high-order information. To reduce the error caused by heatmap discretization, heatmap subpixel regression \cite{Wan2020RobustFA} and subpixel heatmap regression \cite{bulat2021subpixel} methods are proposed respectively for achieving more accurate landmark detection. In LUVLi \cite{Kumar2020LUVLiFA}, a novel end-to-end framework is proposed to solve facial landmark detection problems by jointly estimating the facial landmark locations, uncertainty and visibility, which yields state-of-the-art estimates of the landmark locations. By paying attention to the error-bias with respect to error distribution of facial landmarks, anisotropic direction loss \cite{Huang2021ADNetLE} and anisotropic attention module are proposed to learn more effective facial structures and textures for achieving more accurate landmark detection. Due to the well-designed models, the performance of the heatmap regression-based method has been improved. However, the heatmap regression-based method is actually a dense prediction task, in which a probability is needed to be estimated for each element. It is very hard to accurately estimate the values of all elements in landmark heatmaps and maintain their original distributions. Moreover, their feature learning and regression analysis capabilities also limit the detection accuracy. 

To address the above problem, we propose a Reference Heatmap Transformer (RHT) which enables our method to produce more accurate landmark heatmaps and achieve more precise facial landmark detection by transforming the reference heatmap information. 
\section{Precise Facial Landmark Detection by Reference Heatmap Transformer}
In this section, we introduce the proposed Precise Facial Landmark Detection by Reference Heatmap Transformer. The Reference Heatmap Transformer (RHT) and the Multi-Scale Feature Fusion Module (MSFFM) will be discussed in \textbf{Section 3.1} and \textbf{3.2}. Then, we elaborate the objective function in \textbf{Section 3.3}. Finally, \textbf{Section 3.4} presents the implementation details.
\subsection{Reference Heatmap Transformer}
Heatmap Regression-based FLD methods\cite{Yang2017StackedHN, Dong2018StyleAN, Liu2019SemanticAF} have got promising results as they can effectively encode the part constraints and preserve the spatial relationship between pixels. However, they are still suffering from faces with large poses and heavy occlusions, as in these cases they fail to learn effective feature representations and facial shape constraints. In addition to modeling facial shape constraints only from the original image, the proposed RHT aims to enhance this process by introducing reference heatmap information. By transforming the reference heatmap information into the target landmark heatmaps, the prior knowledge from reference heatmap information can be better preserved and used to model target facial shape constraints, especially for faces with large poses and heavy occlusions. Therefore, the robustness of our proposed method has been enhanced. The proposed RHT includes a Soft Transformation Module (STM) and a Hard Transformation Module (HTM). STM can dynamically select the correlated reference heatmap features to construct the target heatmap feature, and HTM constructs the target heatmap feature by transforming the reference heatmap feature with a 2D affine transformation. They cooperate with each other to model more accurate facial shape constraints. Details will be discussed below.

\textbf{Soft Transformation Module. }STM aims to transform the reference heatmap information (i.e., feature maps) into the target one by estimating the correlation of pixels between the target image and the reference image. As shown in Fig. \ref{fig2}, STM takes the local feature descriptors learned from the target image $I_{tar}$, reference image $I_{ref}$ and reference heatmap $O_{ref}$ as inputs, denoted by ${F_Q}$, ${F_K}$ and ${F_V}$, respectively. To obtain local feature descriptors, we design a local feature extractor, which contains several convolution, batch-norm and Relu operations. The learned local feature descriptors are then treated as three basic elements (${F_Q}$ (query), ${F_K}$ (key), and ${F_V}$ (value)) of the attention mechanism inside a transformer for correlation estimation. To be specific, STM first unfolds both ${F_Q}$, ${F_K}$ into patches, i.e., ${f^i_Q}$ and ${f^j_K}$, in which $i,j \in \left[ {1,H \times W} \right]$, $W$ and $H$ denote the height and width of the feature maps corresponding to the local feature descriptors. Then, the correlation between paired patches ${f^i_Q}$ and ${f^j_K}$ are calculated as their inner product. Note that, before calculating the correlations, each patch should be followed by a $l_2$ normalization. Correlations between all paired patches form a correaltion matrix (denoted as $\bf{C}$) and all the most correlated locations form a index matrix (denoted as $\bf{D}$). Then, for each query $f^i_Q$, the index matrix $\bf{D}$ is used to select the most correlated features in $F_V$ for constructing the transferred features $F_T$. Suppose $f_T^i$ is selected from the ${\bf{D}}^i$-th position of $F_V$, and ${\bf{D}}^i$ can be calculated as follows: 

\begin{small}
	\begin{equation}
		{{\bf{D}}}^i = \mathop {\arg \max }\limits_j {{\bf{C}}^{i,j}}
\end{equation}\end{small}  

So far, the correlated features have been transferred from the reference heatmap information. However, each spatial location of transferred features should be given a different weight as different pairs of patches have different correlations, i.e., highly correlated heatmap transfer should be enhanced while the less correlated ones should be released. To achieve this goal, we construct an attention matrix $\bf{A}$ from ${{\bf{C}}}$ to represent the confidence of the transferred heatmap features for each position in $F_T$. The $i$-th element of the attention matrix is calculated as follows:

\begin{small}
	\begin{equation}
		{\bf{A}}^i = \mathop {\max }\limits_j {{\bf{C}}^{i,j}}
\end{equation}\end{small}where ${\bf{A}}^i$ denotes the $i$-th position of the attention martrix $\bf{A}$. Then, the final transferred features can be formulated as follows:

\begin{small}
	\begin{equation}
		{F_{S}} = {F_T} \odot {\bf{A}}
\end{equation}\end{small}where $\odot$ denotes the element-wise multiplication. With the above index matrix $\bf D$ and attention matrix $\bf A$, STM can effectively transfer the correlated reference heatmap information, i.e., the prior knowledge of reference heatmap information has been introduced. At the same time, compared to the original reference heatmap features, the transferred features usually become discrete due to the difference (e.g., face covers or occlusions) between the two faces. To address this problem, we propose a Hard Transformation Module.

\textbf{Hard Transformation Module. }HTM aims to transform the reference heatmap information into the target ones by utilizing a 2D affine transformation. To be specific, following STN \cite{Jaderberg2015SpatialTN}, HTM first uses a localization network to estimate the affine transformation matrix $\bf{\Theta}$ (i.e., 6 affine transformation parameters) between the reference image and the target one. After that, we need to perform a warping of the input feature maps $F_V$ to obtain the output features maps (i.e., the hard transformation feature maps $F_E$). In general, the output pixels are defined to lie on a regular grid of pixel $(x_i^E,y_i^E)$, and the pointwise transformation can be expressed as follows:

\begin{small}
	\begin{equation}
		\begin{array}{l}
			{\left( {x_i^V,y_i^V} \right)^{\rm{T}}} = {\bf{\Theta} }{\left( {x_i^E,y_i^E,1} \right)^{\rm{T}}}\\
			= \left[ {\begin{array}{*{20}{c}}
					{{\bf{\Theta}} ^{1,1}}&{{\bf{\Theta}} ^{1,2}}&{{\bf{\Theta}} ^{1,3}}\\
					{{\bf{\Theta}} ^{2,1}}&{{\bf{\Theta}} ^{2,2}}&{{\bf{\Theta}} ^{2,3}}
			\end{array}} \right]{\left( {x_i^E,y_i^E,1} \right)^{\rm{T}}}
		\end{array}
\end{equation}\end{small}where $(x_i^E, y_i^E)$ are the target coordinates of the regular grid in the output feature map, $(x_i^V, y_i^V)$ are the source coordinates in the input feature map that define the sample points. Finally, bilinear interpolation is used to obtain the final hard-transformed features $F_E$. By transforming the reference heatmap information with a 2D transformation, their integrity and continuity can be preserved to help enhance the facial shape constraints and produce more accurate target landmark heatmaps. 
\begin{figure*}
	\begin{center}
		\includegraphics[width=0.95\linewidth]{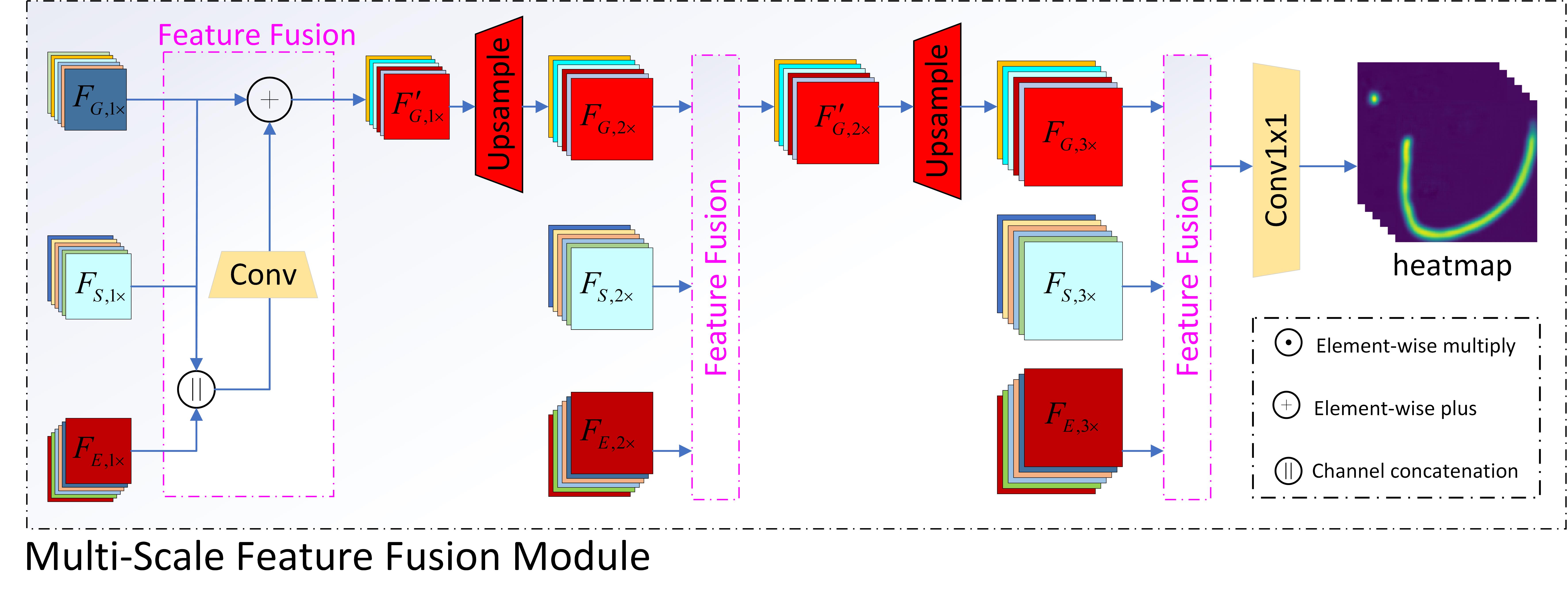}
	\end{center}
	\caption{The architecture of the proposed Multi-Scale Feature Fusion Module (MSFFM). The MSFFM is able to fuse the transformed reference heatmap feature with the semantic features learned from the original face images in a multi-scale way to enhance feature representations and generate more accurate landmark heatmaps.}
	\label{fig3}
\end{figure*}
\subsection{Multi-Scale Feature Fusion Module}
Since STM and HTM transform the reference heatmap information in a soft and hard way, they can cooperate with each other to model more effective facial shape constraints. At the same time, we also use a mainstream backbone to learn the semantic features (denoted as $F_{G}$) from the original image, which will then be fused with the transformed heatmap features $F_E$ and $F_S$ to produce the target heatmaps. Moreover, to make full use of the reference heatmap information, the proposed RHT transforms it in multiple scales (i.e., 1x, 2x, 4x), and a Multi-Scale Feature Fusion Module (MSFFM) is designed to fuse them with $F_{G}$. As shown in Fig. \ref{fig3}, the proposed MSFFM is constructed by using multiple Feature Fusion (FF) blocks, and the process of a FF block can be formulated as follows:

\begin{small}
	\begin{equation}
		\begin{array}{l}
			{{F'}_{G,1 \times }} = {\rm{FF}}\left( {{F_{E,1 \times }},{F_{S,1 \times }},{F_{G,1 \times }}} \right)\\
			= {F_{G,1 \times }} + {\rm{Conv}}\left( {{\rm{Concat}}\left( {{F_{E,1 \times }},{F_{S,1 \times }},{F_{G,1 \times }}} \right)} \right)
		\end{array}
\end{equation}\end{small}
\begin{small}
	\begin{equation}
		{F_{G,2\times}} = {\rm{Dconv}}({F'_{G,1\times}}) 
\end{equation}\end{small}where $\rm{Conv}$, $\rm{Dconv}$ and $\rm{Concat}$ represent a convolution layer, transposed convolution layer and Concatenation operation, respectively. The larger the scale, the closer the corresponding feature is to the real landmark heatmap. By transforming the reference heatmap information at different scales, features at multiple scales can be combined to learn more effective facial shape constraints. Then, with the proposed MSFFM, the transformed multi-scale heatmap feature can be fused with the semantic features learned from the original face images to enhance feature representations and generate more effective landmark heatmaps, thereby achieving more precise facial landmark detection.


\subsection{Objective Function}
The proposed method outputs landmark heatmaps and boundary heatmaps, hence, the loss between the produced heatmaps and ground-truth ones should be used as part of the objective function. Suppose there are $N$ images in the training set, each face contains $M$ landmarks and $P$ boundaries, the objective function can be formulated as follows: 

\begin{small}
	\begin{equation}
		\begin{array}{l}
			\mathbb{L}_1 = \frac{1}{N}\sum\limits_{n = 1}^N \biggl\{  \frac{1}{M}\sum\limits_{m = 1}^M {\left( {{{\gamma}_{n,m}}\left\| {{O_{n,m}} - O_{n,m}^ * } \right\|_2^2} \right)} \\
			{\kern 1pt}{\kern 1pt}{\kern 1pt}{\kern 1pt}{\kern 1pt}{\kern 1pt}{\kern 1pt}{\kern 1pt} + \frac{1}{P}\sum\limits_{p = 1}^P {\left( {\left\| {{B_{n,p}} - B_{n,p}^ * } \right\|_2^2} \right)} \biggr\}
		\end{array}
\end{equation}\end{small}where ${\gamma}_{n, m}$ indicates whether the landmark is visible or not. ${\gamma}_{n, m}=1$ indicates the landmark is visible, and ${\gamma}_{n, m}=0$ indicates invisible. $O$ and $B$ denote the produced landmark heatmaps and boundary heatmaps respectively, and $O^*$ and $B^*$ denote the ground-truth ones.

To perform more effective affine transformations, we design a new loss by using the loss of output feature maps between HTM and STM as a part of the objective function and the detailed loss function is defined as follows:

\begin{small}
	\begin{equation}
		\mathbb{L}_2 = {\left\| { {{F_E}} \odot {\bf{A}} - {F_S}} \right\|_1}
\end{equation}\end{small}where $\bf A$ is the attention matrix and $\odot$ denotes the element-wise multiplication. The attention matrix $\bf A$ can highlight highly correlated pixels and suppress less correlated ones. Less correlated pixels mean that there are large differences between pixels around a landmark in the target image and the reference image. Therefore, by paying more attention to the highly correlated pixels, $\mathbb{L}_2$ can estimate a more accurate affine transformation matrix. Moreover, the 2D affine transformation can also preserve the less correlated heatmap information. These all ensure more effective preservation and transformation of useful reference heatmap information. Finally, by combining $\mathbb{L}_1$ and $\mathbb{L}_2$, the overall objective function can be defined as follows:

\begin{small}
	\begin{equation}
		\mathbb{L}_{overall} = \mathbb{L}_1 + \lambda \mathbb{L}_2
\end{equation}\end{small}

With the above new objective function, useful reference heatmap information can be transformed to enhance feature representations and model more effective facial shape constraints, thereby achieving more precise facial landmark detection.

\subsection{Implementation Details}
To learn the semantic features from the original image, we use the Resnet-50 \cite{He2016DeepRL} and HRNet-W48 \cite{Sun2019HighResolutionRF} based transformers respectively, which are denoted by \textbf{Trans-R} and \textbf{Trans-H}. The implementation of these two transformers is referred to TransPose \cite{Yang2020TransPoseTE}. Both Trans-R and Trans-H take the original image ($256 \times 256 \times 3$) as inputs, and output $32 \times 32 \times 256$ and $64 \times 64 \times 64$ feature maps, respectively. For the proposed RHT, the inputs are target image ($128 \times 128 \times 3$), reference image ($128 \times 128 \times 3$) and reference heatmap ($128 \times 128 \times (M+P)$), where $M$ and $P$ denote the number of landmarks and boundaries, respectively. The outputs are different scales of transformed heatmap features, i.e., $32 \times 32 \times 256$, $64 \times 64 \times 128$ and $128 \times 128 \times 64$. After that, we combine the transformed heatmap features with the semantic features learned by Trans-R to construct \textbf{RHT-R}, while \textbf{RHT-H} is constructed by fusing the semantic features learned by Trans-H. During training, we randomly sample a training image as the reference image. For testing, we use the most correlated face from the training set as the reference image based on SIFT feature matching \cite{Zhang2019ImageSB, Yang2020LearningTT}. The corresponding comparision results on 300W \cite{Sagonas2016300FI}, COFW \cite{Burgosartizzu2013Robust}, AFLW \cite{Zhu2016UnconstrainedFA} and WFLW \cite{Wu2018LookAB} datasets are shown in Tables \ref{tab300w}--\ref{tabwflw}.

We apply augmentation techniques to the training data by random rotation ($45^\circ$), random scaling ($\pm 20\%$), random crop ($\pm10\%$), random gray (20\%), random blur(30\%), and random occlusion (40\%) and random horizontal flip (50\%). $\lambda$ is hyperparameter to balance the heatmap transformation task and the heatmap generation task, from experimental results, we found $\lambda=0.1$ is a good choice. The training phase takes 150 epochs and the batch size is set to 8. The initial learning rate is $1 \times {10^{{\rm{ - 4}}}}$, and it will be decay to $1 \times {10^{{\rm{ - 5}}}}$ with the cosine annealing learning rate decay. The proposed method is trained with Pytorch on 8 Nvidia Tesla V100 GPUs.
\section{Experiments}
In this section, we first introduce the evaluation settings, including the datasets and evaluation metrics. Then, we compare our method with the state-of-the-art FLD methods on benchmark datasets \cite{Burgosartizzu2013Robust, Sagonas2016300FI, Zhu2016UnconstrainedFA, Wu2018LookAB}. After that, the ablation studies are presented. Finally, we show the experimental results and discussions.

\subsection{Datasets}
We evaluate our proposed method on four challenging datasets, i.e., COFW \cite{Burgosartizzu2013Robust}, 300W \cite{Sagonas2016300FI}, AFLW \cite{Zhu2016UnconstrainedFA} and WFLW \cite{Wu2018LookAB}. Details are shown below.

\textbf{300W \cite{Sagonas2016300FI} }(68 landmarks): 300W training set consists of the training sets of AFW \cite{Belhumeur2011LocalizingPO}, LFPW \cite{Zhu2012FaceDP} and Helen \cite{le2012interactive}, and it contains 3148 face pictures in total. 300W testing set includes IBUG \cite{Sagonas2016300FI} and the testing sets of LFPW and Helen. The 300W testing set is usually divided into three subsets, i.e., Challenging subset, Common subset and Fullset, to more comprehensively verify FLD methods. The Challenging subset (i.e., IBUG) includes 135 more general ``in the wild" images, and the Common subset contains 554 images that are from the LFPW testing set and Helen testing set. The Fullset is composed of the former two subsets.   

\textbf{COFW \cite{Burgosartizzu2013Robust} }(29 landmarks): It is a very challenging dataset on occlusion issues, with 1345 training images and 507 testing images. 845 images of the training set come from the LFPW \cite{Zhu2012FaceDP} and the other images are heavily occluded. The testing set contains 507 face images with heavy occlusions and variations on head poses and facial expressions.

\textbf{AFLW \cite{Zhu2016UnconstrainedFA} }(19 landmarks): The AFLW trainning set contains 20000 images, which has extremely large variations on head poses, i.e., jaw angles ranging from ${\rm{ - }}{120^ \circ }$ to ${\rm{ + }}{120^ \circ }$ and pitching angles ranging from ${\rm{ - }}{90^ \circ }$ to ${\rm{ + }}{90^ \circ }$. The AFLW-full testing set contains 4386 images, in which 1165 images (i.e., AFLW-frontal) are selected to evaluate the landmark detection accuracy on frontal faces.

\textbf{WFLW \cite{Wu2018LookAB} } (98 landmarks): It has 98 landmark annotations and images in WFLW are collected from more complicated scenarios. The training set and testing set include 7500 and 2500 images respectively. WFLW also has other attribute annotations, including occlusion, pose, makeup, lighting, blur, and expression, which can comprehensively evaluate existing FLD methods.

\subsection{Evaluation Metrics}
We evaluate the facial landmark detection results with three evaluation metrics. They are Normalized Mean Error (NME) \cite{Wan2021RobustFL, Wan2020RobustFA},  Area Under the Curve (AUC) \cite{Kumar2020LUVLiFA, Wan2020RobustFA} and Failure Rate \cite{Chen2019FaceAW, Tang2020TowardsEU}. Generally, the NME is defined as follows:

\begin{small}
	\begin{equation}
		{\rm{NME}} = \frac{1}{M }\sum\limits_{m  = 1}^M  {\frac{{{{\left\| {\left( {{{ \alpha}_m },{{\beta}_m }} \right) - \left( {{{\hat \alpha}_m },{{\hat \beta}_m }} \right)} \right\|}_2}}}{d}}
\end{equation}\end{small}where $m$ denotes the landmark index, $(\alpha, \beta)$ and $(\hat \alpha, \hat \beta)$ represent the predicted and ground-truth landmark coordinates respectively. $d$ denotes the normalization term, which can be set to the interpupil distances (i.e., $\rm NME_{ip}$ \cite{Wan2021RobustFL, Wan2020RobustFA}), the interocular distance (i.e., $\rm NME_{io}$ \cite{Sagonas2016300FI, Tang2020TowardsEU}), the geometric mean of the width and height of the ground-truth bounding box (i.e., $\rm NME_{box}$ \cite{Bulat2017HowFA, Zafeiriou2017TheMF}) and the diagonal of the tight bounding box (i.e., $\rm NME_{diag}$ \cite{Sun2019HighResolutionRF, Wu2018LookAB}), respectively. To compute AUC, we firstly plot the cumulative distribution of the fraction of test images whose NME (\%) is less than or equal to the value on the horizontal axis. Then the AUC is computed as the area under that curve. FR is defined by the percentage of testing images whose NME is larger than a certain threshold, which is set to 0.1.

\begin{table}
	\caption{Comparisons with state-of-the-art methods on 300W dataset. $\rm NME_{io}$ and $\rm NME_{ip}$ comparisions are given. (\% omitted).}
	\begin{center}
		\begin{tabular}{p{3.3cm}|p{1.cm}p{1.5cm}p{0.8cm}}
			\hline
			Method  & 
			\begin{tabular}[c]{@{}c@{}}Common\\ Subset\end{tabular} & \begin{tabular}[c]{@{}c@{}}Challenging\\ Subset\end{tabular} & Fullset \\ \hline      
				$\rm NME_{io}$ comparisions
			\\ \hline                                                                    
			PCD-CNN{$\rm _{CVPR18}$}\cite{Kumar2018Disentangling3P} & 3.67 & 7.62      & 4.44    \\
			SAN{$\rm _{CVPR18}$}\cite{Dong2018StyleAN}           & 3.34    & 6.60      & 3.98    \\
			AVS{$\rm _{ICCV19}$}\cite{Qian2019AggregationVS} & 3.21    & 6.49      & 3.86    \\
			LAB{$\rm _{CVPR18}$}\cite{Wu2018LookAB}            & 2.98    & 5.19      & 3.49    \\
			Techer{$\rm _{ICCV19}$}\cite{Dong2019TeacherSS}  & 2.91    & 5.91      & 3.49    \\
			DU-Net{$\rm _{ECCV18}$}\cite{Tang2018QuantizedDC}    & 2.90    & 5.15      & 3.35    \\
			DeCaFa{$\rm _{ICCV19}$}\cite{Dapogny2019DeCaFADC} & 2.93    & 5.26      & 3.39    \\
			HR-Net{$\rm _{19'}$}\cite{Sun2019HighResolutionRF}               & 2.87    & 5.15      & 3.32    \\
			HG-HSLE{$\rm _{ICCV19}$}\cite{Zou2019LearningRF} & 2.85    & 5.03      & 3.28    \\
			AWing{$\rm _{ICCV19}$}\cite{Wang2019AdaptiveWL}     & 2.72  & 4.52   & 3.07 \\
			LUVLi{$\rm _{CVPR20}$}\cite{Kumar2020LUVLiFA}       & 2.76  & 5.16    & 3.23 \\
			ADNet{$\rm _{ICCV21}$}\cite{Huang2021ADNetLE}       & 2.53  & 4.58    & 2.93 \\
			\hline
			\textbf{RHT-R} & \textbf{2.42}  & \textbf{4.44}    & \textbf{2.82}  \\ 
			\textbf{RHT-H} & \textbf{2.34} & \textbf{4.37}    & \textbf{2.74} \\
			\hline
				$\rm NME_{ip}$ comparisions
			\\ \hline
			Honari et al.{$\rm _{CVPR18}$}\cite{Honari2018ImprovingLL}&4.20 &7.78 &4.90 \\
			SBR{$\rm _{CVPR18}$}\cite{Dong2018SupervisionbyRegistrationAU}&3.28 &7.58 &4.10 \\
			TS{$\rm _{CVPR18}$}\cite{Dong2019TeacherSS}&3.17 &6.41 &3.78 \\
			Liu et al.{$\rm _{CVPR19}$}\cite{Liu2019SemanticAF} &	3.45&	6.38&	4.02\\
			ODN{$\rm _{CVPR19}$}\cite{Zhu2019RobustFL}	&3.56	&6.67&	4.17\\
			STKI{$\rm _{ACM MM20}$}\cite{Zhu2020SpatialTemporalKI}	&3.36	&7.39&	4.16\\
			MMDN{$\rm _{TNNLS21}$}\cite{Wan2021RobustFL}	&3.17	&6.08&	3.74\\\hline
			\textbf{RHT-R} & \textbf{2.87} & \textbf{5.88}    & \textbf{3.46}  \\ 
			\textbf{RHT-H} & \textbf{2.76} & \textbf{5.69}    & \textbf{3.33}  \\ 
			\hline
		\end{tabular}
	\end{center}
	\label{tab300w}
\end{table}
\subsection{Evaluation under Normal Circumstances}
For benchmark datasets such as 300W \cite{Sagonas2016300FI}, COFW \cite{Burgosartizzu2013Robust}, AFLW \cite{Zhu2016UnconstrainedFA} and WFLW \cite{Wu2018LookAB}, faces in 300W common subset and AFLW-frontal dataset are closer to neutral faces and have smaller variations on the head pose, facial expression and occlusion. Hence, we evaluate the effectiveness of the proposed method under normal circumstances with these two subsets. 

For 300W common subset, as shown in Table \ref{tab300w}, our RHT-H and RHT-R can achieve 2.34\% and 2.42\% $\rm NME_{io}$, and 2.76\% and 2.87\% $\rm NME_{ip}$. In particular, RHT-H improves the metric by 7.51\% in $\rm NME_{io}$ over the best method ADNet \cite{Huang2021ADNetLE} and by 12.93\% in $\rm NME_{ip}$ over the best method MMDN \cite{Wan2021RobustFL}. Moreover, our proposed RHT-R and RHT-H both outperform the state-of-the-art facial landmark detection methods \cite{Wang2019AdaptiveWL, Zhu2020SpatialTemporalKI, Huang2021ADNetLE, Kumar2020LUVLiFA, Wan2021RobustFL}.

For AFLW-frontal dataset, Table \ref{tabaflw} shows our proposed RHT-R and RHT-H can achieve 1.02\% and 0.97\% $\rm NME_{diag}$, which outperform the state-of-the-art methods \cite{Wu2018LookAB, Sun2019HighResolutionRF, Chen2018KernelDN, Kumar2020LUVLiFA}. 

From the experimental results as shown in Tables \ref{tab300w} and \ref{tabaflw}, we can find that our proposed method achieves large improvements on both 300W Common subset and AFLW-frontal subset, which mainly because 1) the prior knowledge from reference heatmap information can be used to model more effective facial shape constraints, 2) the transformation of reference heatmap information is helpful to enhance feature representations and generate more accurate landmark heatmaps, and 3) by fusing the transformed reference heatmap feature with the original image's feature, the proposed method can achieve more precise facial landmark detection.

\begin{table}
	\caption{Comparisons with state-of-the-art methods on AFLW dataset. $\rm NME$ and $\rm AUC^7_{box}$ comparisions are given.  (\% omitted, - not counted).}
	\begin{center}
		\begin{tabular}{p{2.5cm}|p{0.8cm}p{0.9cm}|p{1.1cm}|p{1.1cm}}
			\hline
			\multirow{2}{*}{Method} & \multicolumn{2}{|c|}{$\rm NME_{diag}$} &{$\rm NME_{box}$} & {$\rm AUC^7_{box}$} \\
			\cline{2-5}
			& Full & Frontal & Full & Full \\
			\cline{2-5}	
			\cline{0-0}
			CCL{$\rm _{CVPR16}$}\cite{Zhu2016UnconstrainedFA} & 2.72 & 2.17 & - & -  \\
			LLL{$\rm _{ICCV19}$}\cite{Robinson2019LaplaceLL} & 1.97 & - & - & -  \\
			SAN{$\rm _{CVPR18}$}\cite{Dong2018StyleAN} & 1.91 & 1.85 & - & -  \\
			DSRN{$\rm _{CVPR18}$}\cite{Miao2018DirectSR} & 1.86 & - & - & -  \\	
			LAB{$\rm _{CVPR18}$}\cite{Wu2018LookAB} & 1.85 & 1.62 & - & -  \\
			HR-Net{$\rm _{19'}$}\cite{Sun2019HighResolutionRF} & 1.57 & 1.46 & - & -  \\
			Wing{$\rm _{CVPR17}$}\cite{Feng2017WingLF} & - & - & 3.56 & 5.35  \\ 
			KDN\cite{Chen2018KernelDN} & - & - & 2.80 & 60.3  \\ 
			LUVLi{$\rm _{CVPR20}$}\cite{Kumar2020LUVLiFA} & 1.39 & 1.19 & 2.28 & 68.0  \\ \hline
			\textbf{RHT-R} & \textbf{1.18} & \textbf{1.02} & \textbf{1.99} & \textbf{69.8}  \\ 
			\textbf{RHT-H} & \textbf{1.12} & \textbf{0.97} & \textbf{1.87} & \textbf{70.6} \\ \hline
		\end{tabular}
	\end{center}
	\label{tabaflw}
\end{table}
\begin{table}
	\caption{Comparisons ($\rm NME_{box}$) with state-of-the-art methods on COFW-29 dataset. (\% omitted, - not counted). }
	\begin{center}
		\begin{tabular}{p{3.5cm}|p{1.2cm}p{1.2cm}|p{1.2cm}}			\hline
			Method & NME & $FR_{10\%}$ & $AUC^7_{10\%}$ \\
			\hline
			Human\cite{Burgosartizzu2013Robust} &5.60  &- &-\\
			PCPR\cite{Burgosartizzu2013Robust} &8.50  &20.00 &-\\
			TCDCN\cite{zhang2014facial} &8.05  &- &-\\
			DAC\_CSR\cite{Feng2017DynamicAC} &6.03  &4.73 &-\\
			Wu et al.\cite{Wu2015RobustFL} &5.93  &- &-\\
			Wing{$\rm _{CVPR17}$}\cite{Feng2017WingLF} &5.44  &3.75 &-\\	
			DCFE{$\rm _{ECCV18}$}\cite{Valle2018ADC} &5.27  &- &35.86\\
			LAB{$\rm _{CVPR18}$}\cite{Wu2018LookAB} &5.58  &2.76 &-\\
			ODN{$\rm _{CVPR19}$}\cite{Zhu2019RobustFL}&5.30 &- & - \\
			Awing{$\rm _{ICCV19}$}\cite{Wang2019AdaptiveWL} &4.94  &0.99 &64.40\\
			ADNet{$\rm _{ICCV21}$} \cite{Huang2021ADNetLE} &4.68  &0.59 &53.17\\	
			\hline
			\textbf{RHT-R} & \textbf{4.42} & \textbf{0.79} & \textbf{57.96}   \\
			\textbf{RHT-H} & \textbf{4.38} & \textbf{0.59} & \textbf{59.10}  \\ \hline
		\end{tabular}
	\end{center}
	\label{tabcofw}
\end{table}

\begin{table*}
	\caption{$\rm NME_{io}$ comparisons on WFLW dataset. (\% omitted, - not counted).}
	\begin{center}
		\begin{tabular}{p{3cm}p{1.3cm}p{1.3cm}p{1.5cm}p{1.7cm}p{1.5cm}p{1.3cm}p{1.2cm}}
			\hline
			Method  &
			\begin{tabular}[l]{@{}l@{}}Testset\\\end{tabular} & \begin{tabular}[l]{@{}l@{}}Pose\\ Subset\end{tabular} &
			\begin{tabular}[c]{@{}l@{}}Expression\\ Subset\end{tabular} &
			\begin{tabular}[c]{@{}l@{}}Illumination\\ Subset\end{tabular} &
			\begin{tabular}[c]{@{}l@{}}Make-Up\\ Subset\end{tabular} &
			\begin{tabular}[c]{@{}l@{}}Occlusion\\ Subset\end{tabular} &
			\begin{tabular}[c]{@{}l@{}}Blur\\ Subset\end{tabular}  \\ \hline
			CCFS{$\rm _{CVPR15}$}\cite{zhu2015face} &9.07 &21.36 &10.09 &8.30 &8.74 &11.76 &9.96 \\
			DVLN{$\rm _{CVPR17}$}\cite{wu2017leveraging} &6.08 &11.54 &6.78 &5.73 &5.98 &7.33 &6.88 \\
			LAB{$\rm _{CVPR18}$}\cite{Wu2018LookAB} &5.27 &10.24 &5.51 &5.23 &5.15 &6.79 &6.32 \\
			Wing{$\rm _{CVPR18}$}\cite{Feng2017WingLF} &5.11 &8.75 &5.36 &4.93 &5.41 &6.37 &5.81 \\
			MHHN{$\rm _{TIP21}$}\cite{Wan2020RobustFA} &4.77 &9.31 &4.79 &4.72 &4.59 &6.17 &5.82 \\
			MMDN{$\rm _{TNNLS21}$}\cite{Wan2021RobustFL}	&4.87 &8.15 &4.99 &4.61 &4.72 &6.17 &5.72 \\
			HRNet{$\rm _{19'}$}\cite{Sun2019HighResolutionRF} &4.60 &7.86 &4.78 &4.57 &4.26 &5.42 & 5.36 \\
			AWing{$\rm _{ICCV19}$}\cite{Wang2019AdaptiveWL} &4.36 &7.38 &4.58 &4.32 &4.27 &5.19 &4.96 \\
			LUVLi{$\rm _{CVPR20}$}\cite{Kumar2020LUVLiFA} &4.37 &7.56 &4.77 &4.30 &4.33 &5.29 &4.94 \\
			SCPAN{$\rm _{TCYB22}$} \cite{Wan2021RobustAP} &4.29 &7.22  &4.68 &4.34 &4.21 &5.25 &4.88\\
			ADNet{$\rm _{ICCV21}$} \cite{Huang2021ADNetLE} &4.14 &-  &- &- &- &- &-\\
			\hline
			\textbf{RHT-R} & \textbf{4.01} & \textbf{6.87} & \textbf{4.45 } & \textbf{4.05 } & \textbf{4.11 } & \textbf{4.98 } & \textbf{4.69} \\ 
			\textbf{RHT-H} & \textbf{3.96} & \textbf{6.77} & \textbf{4.38} & \textbf{4.02} & \textbf{4.03} & \textbf{4.77} & \textbf{4.58} \\
			
			\hline
		\end{tabular}
	\end{center}
	\label{tabwflw}
\end{table*}

\subsection{Evaluation of Robustness against Occlusions}
Variations of occlusion and illumination are classic problems in facial landmark detection. The state-of-the-art facial landmark detection methods still suffer from faces with heavy occlusions and complicated illuminations. In this paper, we use COFW dataset \cite{Burgosartizzu2013Robust}, 300W challenging subset \cite{Sagonas2016300FI} and WFLW dataset \cite{Wu2018LookAB} to evaluate the robustness of the proposed method against occlusions.

For 300W challenging subset, our RHT-H and RHT-R can achieve 4.37\% and 4.44\% $\rm NME_{io}$ as shown in Table \ref{tab300w}, which outperform the state-of-the-art FLD methods \cite{Dong2018StyleAN, Dong2019TeacherSS, Dapogny2019DeCaFADC, Sun2019HighResolutionRF, Wang2019AdaptiveWL, Kumar2020LUVLiFA}. This indicates that the proposed method can effectively improve the landmark detection accuracy for heavily occluded faces.

For COFW dataset, RHT-H can boost the $\rm NME_{box}$ to 4.38\%, the failure rate to 0.59\% and the $\rm AUC^7_{box}$ to 59.10\%, which outperforms the state-of-the-art methods \cite{Feng2017DynamicAC, Kumar2018Disentangling3P, Feng2017WingLF, Wu2018LookAB, Wang2019AdaptiveWL, Zhu2019RobustFL, Wan2020RobustFA}. These results indicate that the proposed method can achieve more precise landmark detection for faces with complicated occlusions and illuminations.

As shown in Table \ref{tabwflw}, we can find that our proposed method outperforms the state-of-the-art methods \cite{Wu2018LookAB, Feng2017WingLF, Dapogny2019DeCaFADC, Qian2019AggregationVS, Wang2019AdaptiveWL, Kumar2020LUVLiFA, Huang2021ADNetLE} on the Illumination subset, the Make-Up Subset and the Occlusion Subset. Moreover, our RHT-R and RHT-H can achieve 4.01\% and 3.96\% $\rm NME_{io}$ on WFLW testset, which surpasses the best method ADNet \cite{Huang2021ADNetLE} by 3.14\% and 4.35\% respectively. These all indicate the effectiveness of the proposed models.

Hence, from the experimental results illustrated in Tables \ref{tab300w}, \ref{tabcofw} and \ref{tabwflw}, we can conclude that by introducing the reference heatmap information, our proposed method is able to utilize more prior knowledge to model more effective facial shape constraints and enhance feature representations. Therefore, more accurate target heatmaps can be generated and the robustness of our method against faces under heavy occlusions and complicated illuminations has been enhanced.
\subsection{Evaluation of Robustness against Large Poses}
Faces with large poses is another huge challenge in facial landmark detection. We conduct experiments on AFLW-full \cite{Zhu2016UnconstrainedFA}, 300W challenging subset \cite{Sagonas2016300FI} and WFLW dataset \cite{Wu2018LookAB} to further evaluate the performance of the proposed method on faces with large poses. 

For AFLW-full dataset, Tables \ref{tabaflw} shows our proposed RHT-H can achieve 1.12\% $\rm  NME_{diag}$, 1.87\% $\rm NME_{box}$ and 70.6\% $\rm AUC_{box}^7$, which exceeds the state-of-the-art methods \cite{Robinson2019LaplaceLL, Dong2018StyleAN, Wu2018LookAB, Sun2019HighResolutionRF, Kumar2020LUVLiFA, Wan2020RobustFA}. In particular, RHT-H improves the $\rm NME_{io}$ metric by 19.42\% over the current best method LUVLi \cite{Kumar2020LUVLiFA}. For 300W dataset, as shown in Table \ref{tab300w}, the $\rm NME_{io}$ and $\rm NME_{ip}$ on 300W challenging subset beat the state-of-the-art facial landmark detection methods \cite{Dong2019TeacherSS, Tang2018QuantizedDC, Dapogny2019DeCaFADC, Zou2019LearningRF, Wang2019AdaptiveWL, Kumar2020LUVLiFA, Huang2021ADNetLE}. For WFLW dataset, our proposed RHT-R and RHT-H can achieve 6.87\% and 6.77\% on WFLW Pose Subset, which also surpass the current FLD methods \cite{Dong2019TeacherSS, Tang2018QuantizedDC, Dapogny2019DeCaFADC, Zou2019LearningRF, Wang2019AdaptiveWL, Kumar2020LUVLiFA, Huang2021ADNetLE}.  

From the above experimental results, we can see that our proposed method is more robust to face under large poses. This indicates the proposed method can 1) utilize the prior knowledge of the reference heatmap information to enhance feature representations to generate more effective landmark heatmaps, and 2) model more effective facial shape constraints to detect more accurate landmarks.  

\subsection{Ablation Studies}
To explore the efficacy and the contribution of each component of our proposed method, we conduct the following comprehensive ablation experiments.

\textbf{Evaluation on different backbones. }The proposed RHT can be integrated with mainstream backbones \cite{He2016DeepRL, Yang2017StackedHN, Sun2019HighResolutionRF}. Here, we conduct experiments by separately using ResNet \cite{He2016DeepRL}, Hourglass network \cite{Yang2017StackedHN} and HRNet \cite{Sun2019HighResolutionRF} to show its superiority. To be specific, the Hourglass networks-based transformer, ResNet-50-based transformer and HRnet-based transformer are separately used. For ResNet-50-based and HRnet-based transformers, we design them according to TransPose \cite{Yang2020TransPoseTE} and they are denoted by \textbf{RHT-R} and \textbf{RHT-H} respectively. For the Hourglass networks-based transformer, we separately use one Hourglass network unit and four Hourglass network units, the corresponding models are denoted by \textbf{RHT-HG} and \textbf{RHT-4HG} respectively. We first use the Hourglass network, ResNet-50, HRNet \cite{Sun2019HighResolutionRF} to learn the 2D spatial structure image feature, which will then be separately flattened into a sequence and input into a transformer encoder. After that, the encoded features and the transformed reference heatmap features are combined by the proposed MSFFM to produce the target landmark heatmaps. Following TransPose \cite{Yang2020TransPoseTE}, 4 encoder layers and 8 heads are used for RHT-R, and 4 encoder layers and 1 head are used for RHT-HG, RHT-4HG and RHT-H.

The corresponding experimental results (i.e., $\rm NME_{io}$ comparisons) on 300W Fullset are shown in Table \ref{tabcom}, from which we can see that all the models are enhanced by equipping them with our proposed reference heatmap transformer and multi-scale feature fusion module. This demonstrates that the proposed reference heatmap transformer can be seamlessly integrated with the mainstream backbones for improving the landmark detection accuracy regardless of network architecture. In addition, we can find that RHT-H with the HRNet-W48 backbone \cite{Sun2019HighResolutionRF} yields the best performance (2.74\% $\rm NME_{io}$ on 300W Fullset) among the five models, which surpasses the baseline by 17.47\%. This further indicates the effectiveness of our proposed reference heatmap transformer and multi-scale feature fusion module.

\begin{table}
	\caption{$\rm NME_{io}$ comparisons with different backbones on 300W Fullset. (\% omitted).}
	\begin{center}
		\begin{tabular}{p{3.0cm}|p{2cm}|p{2cm}}\hline
			Backbone & Baseline  &RHT+MSFFM  \\ \hline
			ResNet-50 &3.43 &2.82 \\
			Sigle HG &3.39 &2.83 \\
			Four HGs &3.38 &2.79 \\
			HRNet-W32 &3.35 &2.77 \\
			HRNet-W48 &3.32 &2.74 \\ \hline
		\end{tabular}
	\end{center}
	\label{tabcom}
\end{table}

\textbf{Evaluation on different reference images. }In general, the performance of reference-based methods highly depends on the similarity between the reference image and the target one. However, for the facial landmark detection task, the human face is the only research object and the deep learning models can embed the facial prior information into themselves during training. Moreover, the proposed STM can effectively select and transfer the correlated heatmap features into the target ones by estimating the correlations between the reference image feature and the target one. Therefore, the negative impact of the similarity between the reference image and the target one can be alleviated. To verify this point, we conduct the following experiments, i.e., the most correlated face (denoted as ${corr}_m$ in Table \ref{tabdif}), the most correlated faces in low-resolution (i.e., $32\times32\times3$ (${corr}_{m1}$) and $64\times64\times3$) (${corr}_{m2}$), the least correlated face (${corr}_{l}$), five randomly selected faces ($mean$) and a neutral face ($neutral$) in the training set are separately selected as the reference images. The low-resolution face is obtained by downsampling the high-resolution image with bicubic degradation, and then we resize it to $128\times128\times3$ for feeding into RHT. The corresponding experimental results are shown in Table \ref{tabdif}, from which we can see that 1) when we choose the most correlated face in the training set as the reference face, the proposed method can obtain the best results, 2) a neutral face can be generally used as the reference face to achieve more precise landmark detection and 3) when we use randomly selected face or the least correlated face or the low-resolution face as the reference one, our proposed RHT-R and RHT-H still outperform the baselines (i.e., Trans-R in Table. \ref{tabablation}). All the observations indicate that 1) the proposed reference heatmap transformer can effectively transform the reference heatmap information into the targe ones and further help improve landmark detection accuracy and 2) the low-resolution reference face will lead to a reduction in accuracy, but the mainstream backbone branch as shown in Fig. \ref{fig2} still help our RHT achieve precise facial landmark detection.

\textbf{Evaluation on cross datasets. }Since 300W (68 landmarks), COFW (29 landmarks), AFLW (19 landmarks), and WFLW (98 landmarks) datasets contain different numbers of landmarks, we additionally introduce the COFW-68 \cite{Burgosartizzu2013Robust} dataset (68 landmarks) to evaluate the performance of our RHT on the 300W dataset for cross-datasets evaluation. Specifically, we train our model on the 300W training set where the reference image is also randomly selected from this training set. But during the testing stage, we choose the most correlated face from the COFW-68 dataset training set as the reference face based on SIFT feature matching, and the target image is also from the 300W testing set (i.e., 300W common subset, challenging subset, and full set). The corresponding experimental results are separately represented by RHT-R* and RHT-H* as shown in Table \ref{tabcross}. Compared to the original RHT-R and RHT-H, RHT-R* and RHT-H* only obtain slightly worse $\rm NME_{io}$ scores. This indicates that 1) choosing the correlated face as the reference one always helps achieve more precise facial landmark detection and 2) the prior knowledge from reference heatmap information is introduced by our RHT for modeling more effective facial shape constraints and detecting more accurate landmarks.

\begin{table}
	\caption{Evaluation on cross dataset, $\rm NME_{io}$ comparision on 300W is given and ``*'' denote the reference image is selected from the COFW-68 \cite{Burgosartizzu2013Robust} training set. (\% omitted).}
	\begin{center}
		\begin{tabular}{p{3.3cm}|p{1.cm}p{1.5cm}p{0.8cm}}
			\hline
			Method  & 
			\begin{tabular}[c]{@{}c@{}}Common\\ Subset\end{tabular} & \begin{tabular}[c]{@{}c@{}}Challenging\\ Subset\end{tabular} & Fullset \\ \hline
			ResNet-50 &2.99 &5.24 &3.43  \\ 
			HRNet-W48 &2.87 &5.15 &3.32 \\
			RHT-R & 2.42  &4.44    &2.82  \\ 
			RHT-H & 2.34 & 4.37   &2.74 \\
			\hline
			RHT-R* & 2.47  &4.50    &2.87  \\ 
			RHT-H* & 2.38 & 4.41   &2.78 \\ \hline
		\end{tabular}
	\end{center}
	\label{tabcross}
\end{table}

\begin{table}
	\caption{$\rm NME_{io}$ comparisons with different reference images on 300W Fullset. ``$corr_m$", ``$corr_{m1}$" and ``$corr_{m2}$" denote the most correlated face in different resolutions (i.e., $128\times128\times3$, $64\times64\times3$ and $32\times32\times3$). ``$corr_l$" denote the least correlated face, $neutral$ denotes the neutral face and $mean$ represents the mean $\rm NME_{io}$ of five randomly selected faces.}
	\begin{center}
		\begin{tabular}{p{0.9cm}|p{0.7cm}|p{1.0cm}|p{0.7cm}|p{0.7cm}|p{0.8cm}|p{0.8cm}}\hline
			Method & $corr_m$ & $neutral$  & $mean$ & $corr_l$ &$corr_{m1}$ & $corr_{m2}$   \\ \hline
			\textbf{RHT-R} &2.82  &2.88 &2.98 &3.04 &2.90 &3.12 \\
			\textbf{RHT-H} &2.74  &2.82 &2.94 &2.99 &2.81 &3.06 \\ \hline
		\end{tabular}
	\end{center}
	\label{tabdif}
\end{table}

\textbf{Evaluation on different modules. }The proposed reference heatmap transformer consists of a Soft Transformation Module (STM) and a Hard Transformation Module (HTM). Moreover, a Multi-scale Feature Fusion Module (MSFFM) is further designed to enhance feature representations. For RHT-R, the baseline is ResNet-50-based transformers (denoted as \textbf{Trans-R}). Then, we separately combine Trans-R with our proposed modules, which are denoted by \textbf{Trans-R + STM}, \textbf{Trans-R + STM + MSFFM}, \textbf{Trans-R + HTM + MSFFM} and \textbf{Trans-R + HTM + STM + MSFFM} (i.e., RHT-R), respectively. Table \ref{tabablation} shows the corresponding experimental results on 300W Fullset \cite{Sagonas2016300FI}, COFW \cite{Burgosartizzu2013Robust} and AFLW-full  \cite{Zhu2016UnconstrainedFA} datasets. When we combine Trans-R with our proposed HTM, STM and MSFFM, the detection accuracy can be improved. This demonstrates that 1) STM can effectively transfer the correlated reference heatmap information into the target one, and further help generate more accurate target heatmaps, 2) By transforming the reference heatmap information with a 2D affine transformation, the integrity and continuity of the original reference heatmap information can be kept for achieving more precise facial landmark detection, 3) HTM and STM can cooperate with each other for more effective reference heatmap information transformation and facial shape constraints modeling, and 4) MSFFM is helpful to fuse the transformed reference heatmap feature and the learned semantic features in a multi-scale way to help produce more accurate landmark heatmaps. In addition, \textbf{Trans-R+STM+MSFFM} surpasses \textbf{Trans-R+HTM+MSFFM} on both 300W Fullset and AFLW-full datasets, but for COFW dataset with heavily occluded faces, \textbf{Trans-R+HTM+MSFFM} beats \textbf{Trans-R+STM+MSFFM}. This indicates that when transforming the reference heatmap information, HTM can effectively keep its integrity and continuity to produce more accurate landmark heatmaps for occluded target faces.
\begin{table}
	\caption{$\rm NME$ comparisons with different modules.  (\% omitted). }
	\begin{center}
		\begin{tabular}{p{3.9cm}|p{1.0cm}|p{1.0cm}|p{1.0cm}}\hline
			Method  &	
			\begin{tabular}[c]{@{}c@{}}300W-\\ Fullset\end{tabular} & COFW &
			\begin{tabular}[c]{@{}c@{}}AFLW-\\ Full	 \end{tabular} \\ \hline 
			\textbf{ResNet-50} 			&3.32 			&5.33 	&1.51  \\
			\textbf{Trans-R} 			&3.21 			&5.17 	&1.43  \\
			\textbf{Trans-R+STM} 		&3.01			&4.81 	&1.30 \\
			\textbf{Trans-R+STM+MSFFM} 	&2.94 			&4.69 	&1.26  \\
			\textbf{Trans-R+HTM+MSFFM} 	&2.98 			&4.65 	&1.29  \\
			\textbf{RHT-R} 			&2.82 			&4.42 	&1.18  \\ \hline
		\end{tabular}
	\end{center}
	\label{tabablation}
\end{table}
\begin{table}
	\caption{Evaluation of boundary loss on 300W, AFLW and COFW datasets, and ``-'' means the corresponding models without boundary loss. (\% omitted). }
	\begin{center}
		\begin{tabular}{p{3.5cm}|p{1.2cm}|p{1.2cm}|p{1.2cm}}\hline
			Method  &	
			\begin{tabular}[c]{@{}c@{}}300W-\\ Fullset\end{tabular} & COFW &
			\begin{tabular}[c]{@{}c@{}}AFLW-\\ Full	 \end{tabular} \\ \hline 
			\textbf{RHT-R} 			&2.82 	&4.42 	&1.18  \\ 	
			\textbf{RHT-H} 			&2.74 	&4.38 	&1.12  \\
			\textbf{RHT-R-} 		&2.89 	&4.55 	&1.23  \\
			\textbf{RHT-H-} 		&2.80 	&4.49 	&1.16  \\
			 \hline
		\end{tabular}
	\end{center}
	\label{tabboundary}
\end{table}

\textbf{Evaluation on boundary information. }The boundary heatmap can help model facial shape constraints and achieve precise FLD. We also conduct the ablation study on boundary information. As shown in Tabel \ref{tabboundary}, when we remove the boundary information, the performance of the corresponding models (i.e., RHT-R- and RHT-H-) is reduced, which demonstrates the effectiveness of boundary information.

\begin{figure*}
	\begin{center}
		\includegraphics[width=0.95\linewidth]{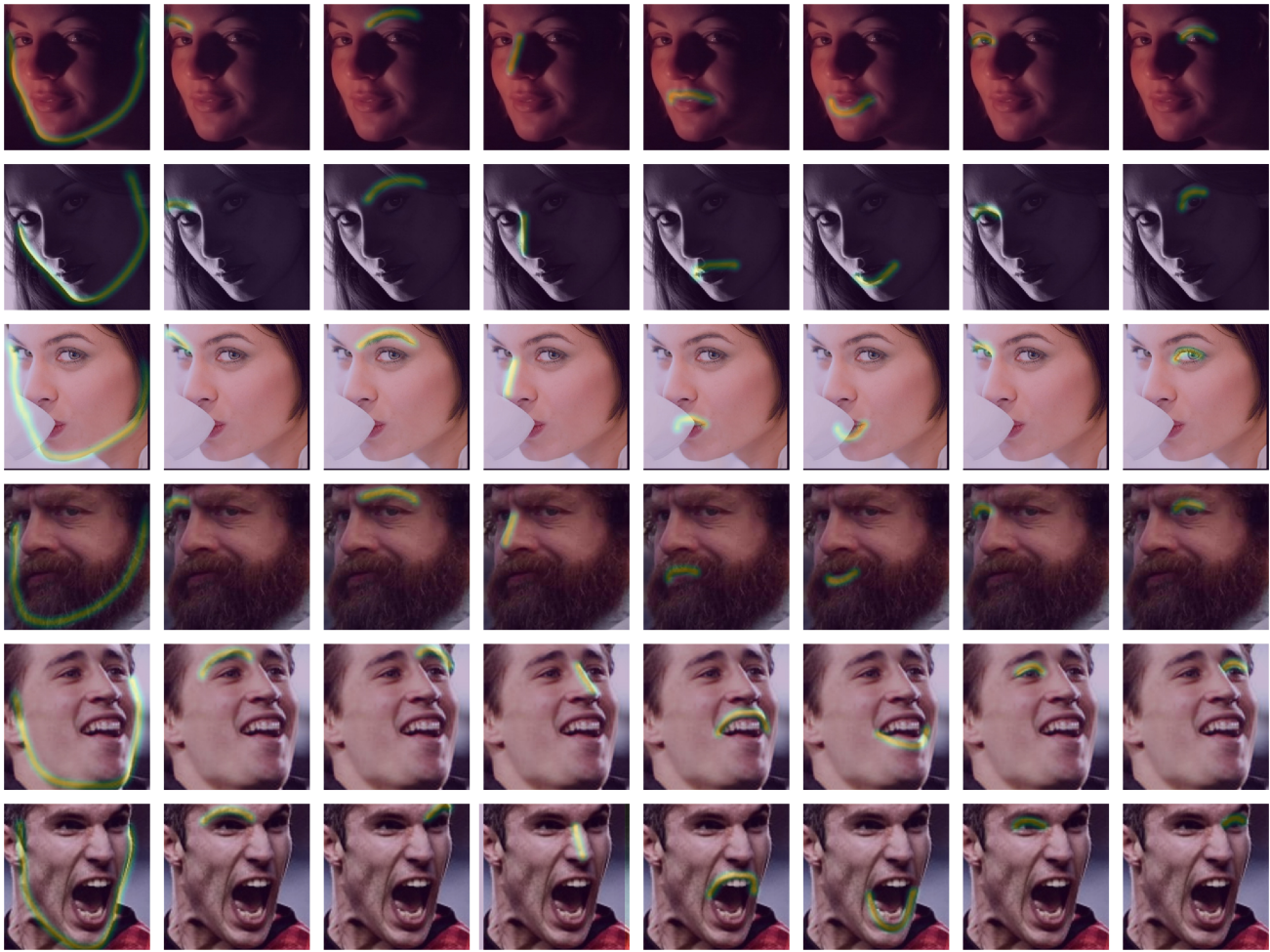}
	\end{center}
	\caption{Visualization of the generated boundary heatmaps. Our proposed method can generate high-quality boundary heatmaps and keep their original distribution for faces with complicated illuminations (the first two rows), heavy occlusions (the middle two rows) and large poses (the last two rows), thereby achieving precise facial landmark detection.}
	\label{fig4}
\end{figure*}

\subsection{Experimental results and discussions}
From the experimental results listed in Tables \ref{tab300w} -- \ref{tabablation} and the figures presented in previous subsections, we state the following observations and corresponding analyses.

(1) Our proposed method, Liu et al. \cite{Liu2019SemanticAF}, MMDN \cite{Wan2021RobustFL}, MHHN \cite{Wan2020RobustFA}, LUVLi \cite{Kumar2020LUVLiFA} and ADNet \cite{Huang2021ADNetLE} are heatmap regression-based facial landmark detection methods. However, from the experimental results on challenging benchmark datasets as shown in Tables \ref{tab300w}-- \ref{tabwflw}, we can find that our proposed method outperforms the other methods, which mainly because 1) the proposed STM can effectively select and transfer the correlated reference heatmap features to help learn more effective facial shape constraints by estimating the correlation between the reference image feature and the target one, 2) The HTM can keep the integrity of reference heatmap information and enhance facial shape constraints by transforming the reference heatmap feature into the target one with a 2D affine transformation, 3) by integrating the STM and HTM into a novel transformer RHT, more prior knowledge from the reference heatmap information can be kept and used to generate more accurate target heatmaps, and 4) by incorporating RHT and MSFFM, the transformed multi-scale reference heatmap feature can be fused with the original image's feature to enhance feature representations and achieve precise facial landmark detection.

(2) Facial shape constraints are very important to precise landmark detection for faces under complicated illuminations and heavy occlusions, our proposed method, LAB \cite{Wu2018LookAB}, MMDN \cite{Wan2021RobustFL}, SCPAN \cite{Wan2021RobustAP} all aim to enhance facial shape constraints by generating facial boundary heatmaps. As shown in Fig. \ref{fig4}, we can find that our proposed method can generate accurate boundary heatmaps for faces with complicated illuminations and heavy occlusions. Moreover, the evaluation results of robustness against occlusions (as shown in \textbf{Section D}) also show the effectiveness of our proposed models. These all indicate that by transforming the reference heatmap information, the prior knowledge of the reference heatmap information can be introduced to achieve more precise facial landmark detection.

(3) The quality of generated landmark heatmaps is critical for precise facial landmark detection, and it is usually more difficult to generate high-quality boundary heatmaps. Hence, we visualize the generated boundary heatmaps in Fig. \ref{fig4}, from which we can find that our proposed method can generate high-quality boundary heatmaps and keep their original distribution for faces with complicated illuminations (the first two rows in Fig. \ref{fig4}), heavy occlusions (the middle two rows in Fig. \ref{fig4}) and large poses (the last two rows in Fig. \ref{fig4}). This indicates that by transforming the reference heatmap information, our proposed method can generate high-quality target heatmaps and achieve more precise facial landmark detection.

(4) Since the variation of faces involves head poses, facial expressions, facial covers and occlusions, it is very hard to choose a good reference face. However, as shown in Table \ref{tabdif}, the detection accuracy drops slightly when we use five randomly selected face images or the neutral face image as the reference ones. Moreover, the experimental results of our proposed method on the Expression Subset of WFLW dataset (as shown in Table \ref{tabwflw}) outperform the other methods. These all indicate that the STM is able to accurately select and transfer the correlated reference heatmap information to help model facial shape constraints and generate landmark heatmaps. Therefore, the impact of the similarity between the reference image and the target one on the detection accuracy can be alleviated by our proposed STM.

(5) Our proposed RHT outperforms the state-of-the-art FLD methods for faces in normal circumstances and large poses. However, for heavily occluded faces (e.g., cup, and beard as shown in Fig. \ref{fig4}), the discriminative information of this occluded area becomes less accurate as it contains a lot of noise. But this situation is alleviated by using the proposed hard transformation module (HTM) and introducing the boundary heatmaps.

\section{Conclusion}
The accuracy of heatmap regression-based FLD methods still encounters great challenges due to inaccurate facial shape constraints modeling and landmark heatmap generation. In this work, we present a reference heatmap transformer to address this problem by transforming the reference heatmap information into the target ones. It is shown that the soft transformation module and hard transformation module designed in RHT can cooperate with each other to model more effective facial shape constraints. Moreover, by incorporating RHT and MSFFM, the transformed heatmap features can be fused with the semantic features learned from the original faces to produce more accurate landmark heatmaps and achieve more precise landmark detection. Experimental results on challenging benchmark datasets demonstrate that the proposed method outperforms the state-of-the-art FLD methods. It can also be found from the experiment that a neutral face with annotated landmarks can be generally used as the reference face of our RHT for improving landmark detection accuracy.

%


\ifCLASSOPTIONcaptionsoff
  \newpage
\fi



%



\bibliographystyle{IEEEtran}
\bibliography{IEEEabrv,IEEEexample}
\end{document}